\documentclass{article}

\usepackage[preprint]{neurips_2026}

\usepackage[utf8]{inputenc} % allow utf-8 input
\usepackage[T1]{fontenc}    % use 8-bit T1 fonts
\usepackage{hyperref}       % hyperlinks
\usepackage{url}            % simple URL typesetting
\usepackage{booktabs}       % professional-quality tables
\usepackage{amsfonts}       % blackboard math symbols
\usepackage{nicefrac}       % compact symbols for 1/2, etc.
\usepackage{microtype}      % microtypography
\usepackage{xcolor}         % colors
\usepackage{multirow}
\usepackage{booktabs}
\usepackage[table]{xcolor}
\usepackage{siunitx}
\usepackage{booktabs}
\usepackage[table]{xcolor}

\usepackage{booktabs}
\usepackage[table]{xcolor}
\usepackage{collcell}
\usepackage{etoolbox}
\usepackage{multirow}
\usepackage{graphicx}
\usepackage{wrapfig}
\usepackage{caption}
\usepackage{amsmath}
\usepackage{subcaption} % 强烈建议使用，NeurIPS 模板兼容性好

\newcommand{\best}[1]{\textbf{#1}}
\newcommand{\second}[1]{\underline{#1}}

\newcommand{\tnum}[1]{\footnotesize #1}

\title{Early High-Frequency Injection for Geometry-Sensitive OOD Detection}

\author{%
  Chuanjie Cheng \\
  Nanjing Normal University
  \And
  Ningkang Peng \\
  Nanjing Normal University
  \And
  Chenxi Liu \\
  Nanjing Normal University
  \And
  Yifan He \\
  Nanjing Normal University
  \And
  Peirong Ma \\
  Nanjing Normal University
  \And
  Yanhui Gu \\
  Nanjing Normal University
}

\begin{document}

\maketitle

\begin{abstract}
Post-hoc OOD detectors score logits or features after training, so their success depends on the geometry already encoded in the representation. We revisit this assumption through a band-wise \(\mathrm{MMD}^2\) analysis across CE, SimCLR, SupCon, and the OOD-oriented representation method PALM. In our diagnostic, low-frequency input bands induce weaker ID/OOD feature discrepancy, whereas higher-frequency bands tend to provide stronger separability. This observation motivates EIHF, an input-side intervention that exposes high-frequency evidence before the first convolution without changing the training objective. EIHF is strongest for geometry-sensitive OOD detection: under matched training and scoring settings, it reshapes class-conditional feature geometry and reduces ID/OOD Mahalanobis score overlap. Experiments on CIFAR-100 and ImageNet-100 show gains on CIFAR-100 and the best average FPR95 with second-best average AUROC on ImageNet-100, while also revealing a limitation on the scene-centric Places shift. Code is available at \url{https://anonymous.4open.science/r/EIHF}.
\end{abstract}

\section{Introduction}

Open-world visual recognition exposes closed-set classifiers to inputs outside their training distribution~\citep{drummond2006openworld,huang2020safety}. Such models can assign high-confidence predictions to out-of-distribution (OOD) samples~\citep{nguyen2015deep,hendrycks2017baseline,liang2018odin}, making OOD detection a central requirement for deployment~\citep{hendrycks2017baseline,yang2024generalized}.

Most OOD detection methods follow a post-hoc scoring paradigm: after training on ID data, they score fixed logits or features. These scores range from confidence- and energy-based criteria to feature-geometric or reconstruction-based measures~\citep{hendrycks2017baseline,liang2018odin,liu2020energy,wang2021energy,lee2018mahalanobis,sun2022knn,wang2022vim,ren2019likelihood,grathwohl2020jem,xiao2020likelihood}, but they commonly assume that the learned representation already provides sufficient ID/OOD structure. We challenge this assumption and argue that OOD detection is also limited by the frequency-dependent geometry of the representation space.

\begin{figure}[t]
    \centering

    \begin{minipage}{0.48\linewidth}
        \centering
        \includegraphics[width=\linewidth]{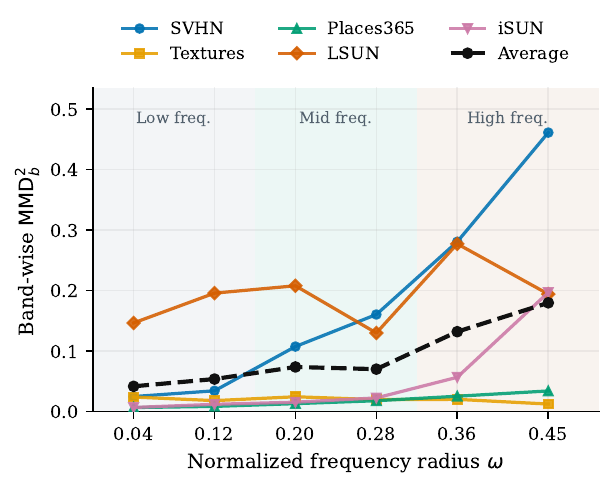}
        \vspace{-2mm}
        \centerline{\small (a) Dataset-wise band-wise \(\mathrm{MMD}^{2}\)}
    \end{minipage}
    \hfill
    \begin{minipage}{0.48\linewidth}
        \centering
        \includegraphics[width=\linewidth]{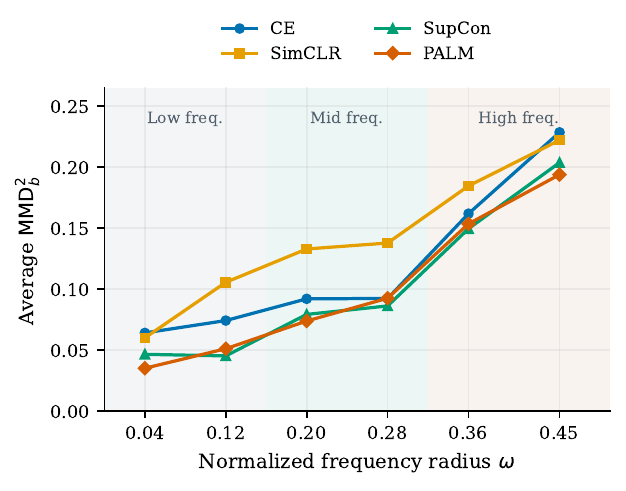}
        \vspace{-2mm}
        \centerline{\small (b) Cross-objective band-wise \(\mathrm{MMD}^{2}\)}
    \end{minipage}
    \vspace{1.5mm}
    \caption{
    Band-wise \(\mathrm{MMD}^{2}\) with CIFAR-100 as ID.
The diagnostic is computed on fixed encoders by feeding band-limited inputs and measuring final-feature discrepancy. Low-frequency bands show weaker ID/OOD feature discrepancy, while mid- and high-frequency bands show stronger separability across OOD datasets and training objectives.
    }
    \label{fig:bandwise_mmd}
\end{figure}
We test this view with a band-wise squared maximum mean discrepancy analysis, or band-wise \(\mathrm{MMD}^2\)~\citep{gretton2012kernel}, under representative training paradigms including CE, SimCLR~\citep{chen2020simclr}, SupCon~\citep{khosla2020supcon}, and PALM~\citep{lu2024palm}. The diagnostic measures the ID/OOD feature discrepancy induced by each input frequency band after encoding. Across OOD datasets and training objectives, we observe a frequency imbalance: low-frequency inputs induce weaker discrepancy, while mid- and high-frequency inputs separate the two distributions more clearly. This observation suggests that fixed post-hoc representations can make frequency components useful for OOD detection unequally accessible. Since low-frequency structure can be shared by ID and OOD samples, post-hoc scores on fixed features may be limited by the frequency-dependent geometry already encoded in the representation.

This diagnosis motivates testing whether early high-frequency exposure can improve the representation used by geometry-sensitive detectors. Such cues can carry OOD information, but they also include within-class variation and sensor noise, so naive amplification may destabilize distance-based ID geometry.

Motivated by this diagnosis, we propose EIHF, an input-side intervention that appends a fixed high-pass residual channel before the first convolution while leaving the training objective unchanged. EIHF changes the representation geometry used by existing detectors without defining a new OOD score, making its effect score-compatible rather than score-uniform. In matched settings, EIHF reshapes class-conditional geometry and enables Mahalanobis-style scoring to better separate ID and OOD samples. Experiments show gains on CIFAR-100, the best average FPR95 among compared methods on ImageNet-100, and strong results on SUN and Textures, but also weaker performance on the scene-centric Places shift.

Our contribution is a diagnosis-to-intervention chain: band-wise separability motivates early high-frequency exposure, and the resulting representation change is evaluated through geometry-sensitive OOD scoring. We summarize:

\begin{itemize}
\item \textbf{Frequency-band diagnosis.}
We use band-wise \(\mathrm{MMD}^2\) to show that, in the tested settings, mid- and high-frequency input bands induce stronger ID/OOD feature discrepancy than low-frequency bands across datasets and training objectives.

\item \textbf{Input-side high-frequency injection.}
We propose EIHF, which appends a fixed high-frequency residual channel before the first convolution without changing the training loss or OOD scoring rule.

\item \textbf{Geometry-centered analysis.}
We show that EIHF improves Mahalanobis-style OOD detection by reducing ID/OOD score overlap, while compactness alone does not explain the observed gains.
\end{itemize}

\section{Related Work}

\subsection{Out-of-distribution Detection}

Out-of-distribution (OOD) detection aims to identify test samples outside the training distribution. Many methods follow a post-hoc scoring paradigm, where a classifier is first trained on ID data and an OOD score is then computed from logits, features, gradients, or reconstruction behavior. Representative examples include confidence- and energy-based scores~\citep{hendrycks2017baseline,liang2018odin,liu2020energy,wang2021energy,wei2022logitnorm}, feature-geometric scores such as Mahalanobis distance and KNN~\citep{lee2018mahalanobis,sehwag2021ssd,sun2022knn}, generative or density-based scores~\citep{kingma2018glow,grathwohl2020jem,ren2019likelihood,xiao2020likelihood}, and activation- or feature-shaping criteria~\citep{sun2021react,djurisic2023ash,wang2022vim,du2022vos,huang2021gradients}. Another line of work improves the representation itself through contrastive learning~\citep{wu2018instance,oord2018cpc,he2020moco,grill2020byol,robinson2021hard}, nearest-neighbor or prototype-based distance scoring~\citep{snell2017prototypical,caron2018deep,caron2020swav,li2021pcl}, and class-conditional structure modeling, including CSI, SSD, KNN+, NPOS, CIDER, PALM, DPL, and DRL~\citep{tack2020csi,sehwag2021ssd,sun2022knn,tao2023npos,ming2023cider,lu2024palm,peng2025dpl,zhang2024drl}. Our work follows this representation-side view, but focuses on frequency-wise ID/OOD separability and uses a lightweight input-side intervention rather than a new score or auxiliary OOD data.

\subsection{Frequency Structure of Learned Representations}

Neural networks exhibit frequency-dependent learning behavior. Studies on spectral bias and the frequency principle show that deep models tend to fit low-frequency components earlier and more easily than high-frequency components~\citep{rahaman2019spectral,xu2020frequency}. In vision, frequency-domain analyses have been used to study classification behavior, texture sensitivity, corruption robustness, adversarial vulnerability, and architectural or objective-induced spectral preferences~\citep{geirhos2019imagenet,jo2017measuring,yin2019fourier,cai2023frequency}. These studies mainly concern classification or robustness. OOD detection raises a different question: how do input frequency bands affect the geometry between ID and OOD feature distributions? We address this question with band-wise \(\mathrm{MMD}^2\).

\subsection{Input-Level High-Pass Augmentation}

Input-level high-pass or edge-aware augmentation has been explored in vision tasks using edge operators, image gradients, high-pass filters, wavelet decompositions, or Fourier-domain transforms as additional channels or auxiliary branches~\citep{canny1986computational,mallat1989theory,bruna2013invariant,yin2019fourier}. These designs are typically motivated by recognition accuracy, robustness, or domain generalization. EIHF is formally related to this line of work, but its motivation and analysis are different. Rather than treating high-frequency maps as generic visual cues, EIHF is derived from a frequency-wise diagnosis of ID/OOD representation separability and is analyzed through its effect on class-conditional geometry and Mahalanobis-style score separation.

\section{Method}
\label{sec:method}
We present EIHF as a diagnosis-driven input-level intervention for OOD detection.
EIHF does not add a high-frequency map as a generic visual cue; it addresses a
frequency-dependent weakness in the representation geometry used by
geometry-sensitive OOD scores. We first define the scoring setting and the
band-wise diagnostic before describing EIHF and its geometric effect.

\begin{figure}[t]
  \centering
  \includegraphics[width=\linewidth,trim=0 150 0 60,clip]{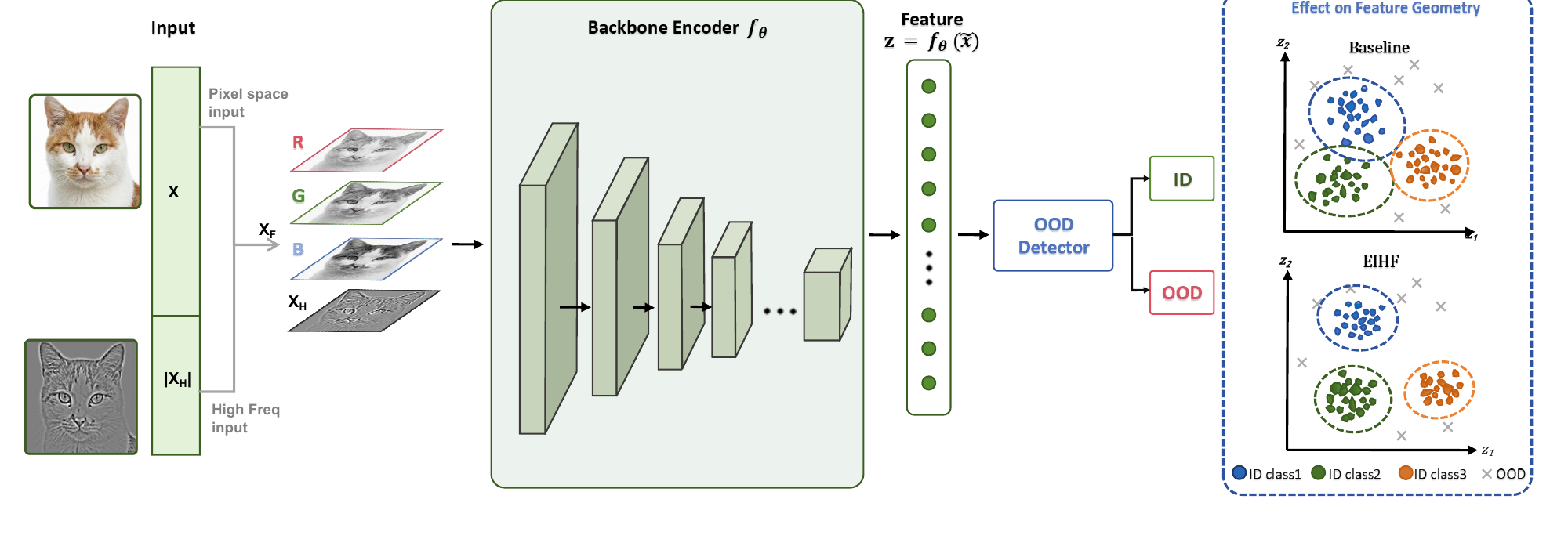}
  \caption{Overview of the EIHF framework. EIHF computes a fixed high-frequency residual from the normalized RGB input, appends it as a fourth channel before the first convolution, and trains the classifier with the same objective. At test time, standard post-hoc OOD scores are applied to the four-channel input, with feature statistics or retrieval banks computed in the same transformed ID space.}
  \label{fig:eihf_framework}
\end{figure}

\subsection{Preliminary}

Let \(\mathcal{D}_{\mathrm{train}}=\{(x_i,y_i)\}_{i=1}^{N}\) denote the ID training set, where \(x_i\in\mathbb{R}^{3\times H\times W}\) and \(y_i\in\{1,\ldots,C\}\). At test time, a sample may come from either the ID distribution \(P_{\mathrm{in}}\) or an unseen OOD distribution \(P_{\mathrm{out}}\). Given a classifier \(F_\theta=g_\theta\circ f_\theta\), most post-hoc detectors assign an ID score from its logits or features:
\begin{equation}
s(x)=\mathcal{S}(x;F_{\theta},\mathcal{D}_{\mathrm{train}}),
\end{equation}
where larger values indicate stronger ID confidence and samples with \(s(x)<\tau\) are rejected as OOD.

To describe input-side interventions, we introduce an input transformation \(\eta\). Standard RGB training corresponds to \(\eta_0(x)=x\). For any transformation \(\eta\), the corresponding training set is
\begin{equation}
\mathcal{D}_{\mathrm{train}}^{\eta}
=
\{(\eta(x_i),y_i)\mid (x_i,y_i)\in\mathcal{D}_{\mathrm{train}}\}.
\end{equation}
This notation matters for feature-based detectors: their class statistics or retrieval banks must be computed in the same transformed input space as the test samples. EIHF instantiates \(\eta\) by augmenting the RGB image with an additional high-frequency residual channel.

\subsection{Diagnostic: Band-wise Feature Separability from Input Frequencies}

Before defining EIHF, we diagnose which input frequency components provide stronger feature-level evidence for ID/OOD separation. Given an RGB baseline encoder \(f_0\) trained on original three-channel inputs, we use it only for this diagnostic analysis and examine how frequency-restricted images are mapped into the final feature space.

Let \(\mathcal{F}\) and \(\mathcal{F}^{-1}\) denote the two-dimensional Fourier transform and its inverse. We partition the normalized frequency radius into \(B\) non-overlapping annular bands, with binary mask \(M_b\) for the \(b\)-th band. The corresponding band-limited image is
\begin{equation}
x^{(b)}
=
T_b(x)
=
\mathcal{F}^{-1}
\left(
M_b \odot \mathcal{F}(x)
\right),
\qquad b=1,\ldots,B,
\end{equation}
where \(\odot\) denotes element-wise multiplication. We apply the same preprocessing and feature extractor across all bands, using the real-valued inverse transform as the network input. We then feed \(x^{(b)}\) into the same encoder and obtain \(z^{(b)}=f_0(x^{(b)})\). Thus, this diagnostic measures how each input frequency subspace is mapped into the learned feature space, rather than the Fourier spectrum of intermediate feature maps.

Given ID samples \(\{x_i\}_{i=1}^{n}\) and OOD samples \(\{o_j\}_{j=1}^{m}\), we collect the corresponding band-induced features
\(\mathcal{Z}_{\mathrm{in}}^{(b)}=\{f_0(T_b(x_i))\}_{i=1}^{n}\) and
\(\mathcal{Z}_{\mathrm{out}}^{(b)}=\{f_0(T_b(o_j))\}_{j=1}^{m}\).
The empirical band-wise discrepancy is
\begin{equation}
\widehat{\mathrm{MMD}}_{b}^{2}
=
\widehat{\mathrm{MMD}}^{2}
\left(
\mathcal{Z}_{\mathrm{in}}^{(b)},
\mathcal{Z}_{\mathrm{out}}^{(b)}; k
\right),
\end{equation}
where \(k\) is an RBF kernel and \(\widehat{\mathrm{MMD}}^{2}\) denotes the empirical squared MMD estimator. For each comparison, the RBF bandwidth is fixed before comparing bands, and the same ID/OOD sample protocol is used for every band and training objective. The resulting curve \(\{\widehat{\mathrm{MMD}}_{b}^{2}\}_{b=1}^{B}\) defines a frequency-wise separability profile, which identifies the input bands that induce stronger ID/OOD feature discrepancy.

This diagnostic is not intended to prove that high-frequency cues are universally beneficial, nor does it by itself establish a causal mechanism for EIHF. Rather, it identifies which input frequency bands induce stronger ID/OOD discrepancy after encoding in the tested settings, with larger \(\widehat{\mathrm{MMD}}_b^2\) indicating stronger band-induced feature separability. We use this observation as motivation for testing an input-side intervention that exposes such evidence early while preserving the original RGB signal.

\subsection{EIHF: Early High-Frequency Injection}

Guided by the band-wise diagnostic, we instantiate the input transformation \(\eta\) as \textbf{Early High-Frequency Injection} (EIHF). EIHF preserves the original RGB input and appends a deterministic residual channel that exposes local high-frequency evidence before the first convolution.

Given a preprocessed RGB tensor \(x\in\mathbb{R}^{3\times H\times W}\), we obtain a single-channel intensity map \(G(x)\in\mathbb{R}^{1\times H\times W}\) using a fixed linear projection of the normalized RGB tensor. We estimate its low-frequency component with a fixed, non-learnable smoothing kernel \(K\), and define the high-frequency residual as
\begin{equation}
C_{\mathrm{hf}}(x)
=
\left|
G(x)-K * G(x)
\right|,
\end{equation}
where \(*\) denotes convolution and \(|\cdot|\) is applied element-wise. Unless otherwise specified, \(K\) is a Gaussian smoothing kernel. This residual emphasizes local variations not explained by the smoothed low-frequency component, such as edges and textures.

The additional channel and augmented input are then defined as
\begin{equation}
e(x)
=
\alpha_{\mathrm{hf}} C_{\mathrm{hf}}(x),
\qquad
\tilde{x}
=
\eta_{\mathrm{EIHF}}(x)
=
\operatorname{concat}_{\mathrm{ch}}(x,e(x))
\in
\mathbb{R}^{4\times H\times W},
\end{equation}
where \(\alpha_{\mathrm{hf}}\) aligns the magnitude of the residual channel with the normalized RGB inputs. We compute this coefficient from the ID training set only. Let \(\sigma_{\mathrm{hf}}\) be the standard deviation of all unscaled residual-channel pixels over the ID training set. We set
\begin{equation}
\alpha_{\mathrm{hf}}
=
\frac{1}{\sigma_{\mathrm{hf}}+\epsilon},
\end{equation}
where \(\epsilon\) is a small constant for numerical stability. The same fixed \(\alpha_{\mathrm{hf}}\) is used for all ID and OOD samples within a benchmark, and no OOD data are used to tune it.

EIHF is applied after the standard RGB preprocessing pipeline; the residual is computed from the normalized RGB tensor observed by the network, and the same preprocessing is used for ID and OOD samples. The only architectural change is extending the first convolution from three to four input channels; subsequent layers, the training objective, and downstream OOD scores remain unchanged.

\begin{figure}[t]
  \centering
  \includegraphics[width=\linewidth]{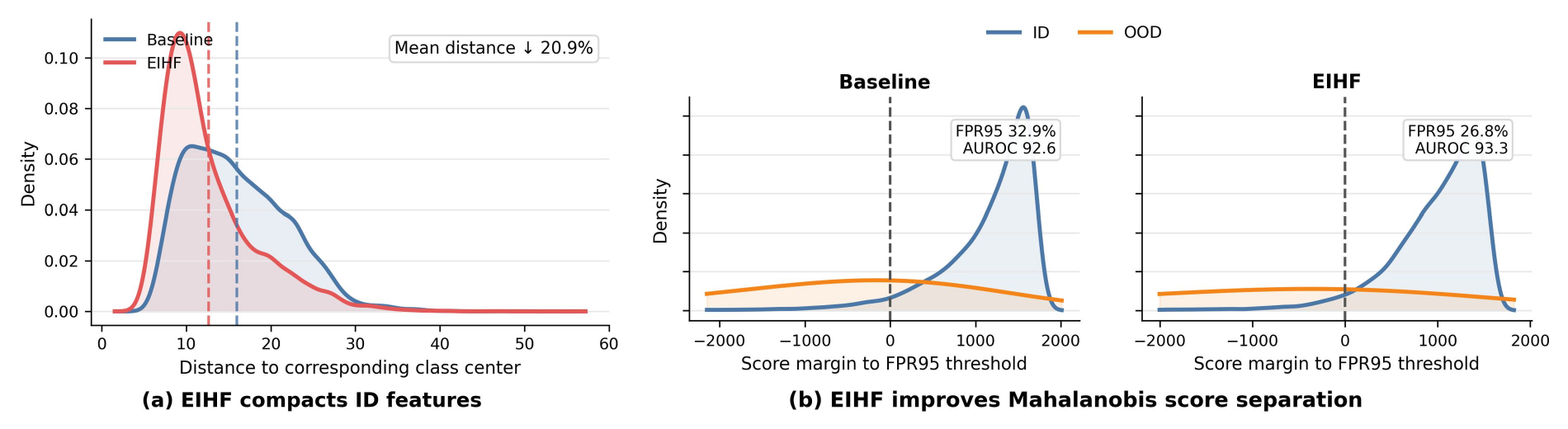}
  \caption{\textbf{Mechanism analysis of how EIHF shapes representation geometry.}
  (a) ID distances to class centers; EIHF reduces the mean distance by 20.9\%.
  (b) ID and OOD Mahalanobis score distributions; EIHF reduces score overlap near the FPR95 operating point. Experiments use ResNet-34 on CIFAR-100.}
  \label{fig:feature_geometry}
\end{figure}

\subsection{OOD Scoring and Geometry Diagnostics}

EIHF changes the input space used for training and post-hoc scoring, but not the scoring rule itself. Training replaces \(x_i\) with \(\tilde{x}_i=\eta_{\mathrm{EIHF}}(x_i)\); at test time,
\[
s_{\mathrm{EIHF}}(x)
=
\mathcal{S}\big(\eta_{\mathrm{EIHF}}(x);F_\theta,\mathcal{D}_{\mathrm{train}}^{\eta_{\mathrm{EIHF}}}\big),
\]
where \(\mathcal{D}_{\mathrm{train}}^{\eta_{\mathrm{EIHF}}}\) is the transformed ID training set. Thus MSP, Energy, Mahalanobis, and KNN can be used unchanged, with the clearest gains expected for geometry-sensitive scores.

\paragraph{Why compactness alone is insufficient.}
A natural ID-geometry diagnostic is the average within-class variance
\[
V_{\mathrm{intra}}
=
\frac{1}{C}
\sum_{c=1}^{C}
\frac{1}{N_c}
\sum_{i:y_i=c}
\left\|
f_\theta(\tilde{x}_i)-\tilde{\mu}_c
\right\|_2^2,
\qquad
\tilde{\mu}_c
=
\frac{1}{N_c}
\sum_{i:y_i=c}
f_\theta(\tilde{x}_i).
\]
With class covariance \(\Sigma_c\), this equals
\[
V_{\mathrm{intra}}
=
\frac{1}{C}
\sum_{c=1}^{C}
\operatorname{tr}(\Sigma_c).
\]
Mahalanobis scoring, however, reweights feature directions by the detector covariance:
\[
d_c(z)
=
(z-\mu_c)^\top
\hat{\Sigma}^{-1}
(z-\mu_c),
\qquad
\mathbb{E}[d_c(z)]
=
\operatorname{tr}(\hat{\Sigma}^{-1}\Sigma_c),
\]
where \(\hat{\Sigma}\) is the detector covariance estimate and the expectation assumes \(z\) is an ID feature from class \(c\). Thus, reducing \(\operatorname{tr}(\Sigma_c)\) alone need not reduce Mahalanobis score overlap. ID compactness is informative, but insufficient to explain OOD performance.

\paragraph{Score overlap and FPR95.}
Let \(s(z)\) be an ID score, where larger values indicate stronger ID confidence. The FPR95 threshold \(\tau_{95}\) and false positive rate are
\[
\Pr_{z\sim P_{\mathrm{in}}}(s(z)\ge \tau_{95})=0.95,
\qquad
\mathrm{FPR95}
=
\Pr_{z\sim P_{\mathrm{out}}}(s(z)\ge \tau_{95})
=
\int_{\tau_{95}}^{\infty} p_{\mathrm{out}}(s)\,ds.
\]
FPR95 is therefore controlled by how much OOD score mass enters the high-ID-confidence region, motivating a direct score-overlap diagnostic.

We estimate overlap by histogramming ID and OOD Mahalanobis scores with shared bins:
\[
\operatorname{Overlap}
=
\sum_{b}
\min\big(p_{\mathrm{in}}(b),p_{\mathrm{out}}(b)\big).
\]
Lower values indicate better score-level separation.

\paragraph{Geometry diagnostics beyond compactness.}
The full diagnostics are reported in Table~\ref{tab:geometry_diagnostic}. EIHF reduces the mean ID distance and achieves the lowest Mahalanobis score overlap, matching its best OOD performance. The zero-channel control has the smallest \(V_{\mathrm{intra}}\), but its score overlap and FPR95 are much worse, showing that compactness by itself is not sufficient. Figure~\ref{fig:score_overlap_fpr} further shows that overlap aligns more consistently with FPR95, supporting our interpretation that EIHF helps mainly by improving ID/OOD score separation.

\section{Experiments}

We evaluate EIHF on CIFAR-10, CIFAR-100, and ImageNet-100 OOD benchmarks. The main comparison tests whether early high-frequency exposure improves geometry-sensitive OOD detection; the score-sensitivity study checks whether the effect depends on the downstream score; and the ablations separate fourth-channel capacity, frequency content, injection location, and spatial alignment.

\subsection{Experimental Setup}

\paragraph{Datasets.}
We use CIFAR-10, CIFAR-100~\citep{krizhevsky2009learning}, and ImageNet-100~\citep{deng2009imagenet,tian2020cmc} as in-distribution (ID) datasets. For CIFAR experiments, models are trained on the training split and evaluated on the test split, with SVHN~\citep{netzer2011svhn}, LSUN~\citep{yu2015lsun}, iSUN~\citep{xu2015isun}, Places365~\citep{zhou2017places}, and Textures~\citep{cimpoi2014dtd} used as OOD test sets. ImageNet-100 evaluates transfer to a larger-scale recognition setting. All OOD datasets are used only at test time.

\paragraph{Training protocol.}
Unless otherwise specified, EIHF follows the PALM training pipeline~\citep{lu2024palm} with the same objective, optimization schedule, augmentation, and hyperparameters. It only replaces \(x\) with \(\tilde{x}=\eta_{\mathrm{EIHF}}(x)\) and extends the first convolution to four channels; all subsequent layers are unchanged. We use ResNet-18 for CIFAR-10, ResNet-34 for CIFAR-100, and ResNet-50 for ImageNet-100~\citep{he2016resnet}.

\paragraph{Baselines and scoring.}
We compare with representative post-hoc and representation-learning methods, including MSP~\citep{hendrycks2017baseline}, ODIN~\citep{liang2018odin}, Energy~\citep{liu2020energy}, ReAct~\citep{sun2021react}, ASH~\citep{djurisic2023ash}, VIM~\citep{wang2022vim}, VOS~\citep{du2022vos}, SSD+~\citep{sehwag2021ssd}, KNN+~\citep{sun2022knn}, CIDER~\citep{ming2023cider}, PALM~\citep{lu2024palm}, DPL~\citep{peng2025dpl}, and DRL~\citep{zhang2024drl}. The main tables are benchmark comparisons under the listed method protocols; the matched EIHF comparisons keep the PALM training pipeline and Mahalanobis-style scoring fixed while changing only the input representation. Unless otherwise specified, EIHF uses Mahalanobis distance~\citep{lee2018mahalanobis}, since it directly depends on class-conditional geometry. Other post-hoc scores are used to measure score sensitivity rather than detector-uniform gains.

\paragraph{Implementation details.}
Experiments are implemented in PyTorch with mixed-precision training. The added first-layer weights for the high-frequency channel are randomly initialized with the default Kaiming uniform distribution. All main experiments run on a single NVIDIA L20 GPU. EIHF-specific preprocessing details are provided in the appendix; the remaining training settings follow the corresponding baseline protocol.
\subsection{Main Results}

\begin{table*}[!t]
\centering
\caption{OOD detection results on CIFAR-10 with ResNet-18. FPR denotes FPR95. $\uparrow$ indicates larger values are better and $\downarrow$ indicates smaller values are better. Best results are shown in bold and second-best results are underlined.}
\label{tab:cifar10_benchmark}
\footnotesize
\setlength{\tabcolsep}{2.8pt}
\renewcommand{\arraystretch}{1.15}

\begin{tabular*}{\textwidth}{@{\extracolsep{\fill}}lcccccccccccc}
\toprule

& \multicolumn{10}{c}{\textbf{OOD Datasets}} 
& \multicolumn{2}{c}{\textbf{Average}} \\

\textbf{Methods}
& \multicolumn{2}{c}{SVHN}
& \multicolumn{2}{c}{Places365}
& \multicolumn{2}{c}{LSUN}
& \multicolumn{2}{c}{iSUN}
& \multicolumn{2}{c}{Textures}
& \\

\cmidrule(lr){2-13}

& FPR$\downarrow$ & AUROC$\uparrow$
& FPR$\downarrow$ & AUROC$\uparrow$
& FPR$\downarrow$ & AUROC$\uparrow$
& FPR$\downarrow$ & AUROC$\uparrow$
& FPR$\downarrow$ & AUROC$\uparrow$
& FPR$\downarrow$ & AUROC$\uparrow$ \\
\midrule
MSP & 59.66 & 91.25 & 62.46 & 88.64 & 51.93 & 92.73 & 54.57 & 92.12 & 66.45 & 88.50 & 59.01 & 90.65 \\
ODIN & 20.93 & 95.55 & 63.04 & 86.57 & 31.92 & 94.82 & 33.17 & 94.65 & 56.40 & 86.21 & 41.09 & 91.56 \\
Energy & 54.41 & 91.22 & 42.77 & 91.02 & 23.45 & 96.14 & 27.52 & 95.59 & 55.23 & 89.37 & 40.68 & 92.67 \\
ReAct & 48.16 & 92.32 & 37.25 & 93.13 & 18.09 & 96.91 & 20.35 & 95.59 & 96.51 & 47.41 & 34.25 & 94.09 \\
ASH & 28.94 & 94.84 & 27.29 & 91.31 & 9.06 & 98.34 & 21.61 & 95.95 & 35.02 & 93.63 & 27.29 & 94.81 \\
Vim & 24.95 & 95.36 & 63.04 & 86.57 & 7.26 & 98.53 & 33.17 & 94.65 & 56.40 & 86.21 & 36.96 & 92.26 \\
VOS & 15.69 & 96.37 & 37.95 & 91.78 & 27.64 & 93.82 & 30.42 & 94.87 & 32.68 & 93.68 & 28.88 & 94.10 \\
SSD+ & 2.47 & 99.51 & 22.05 & 95.57 & 10.56 & 97.83 & 28.44 & 95.67 & 9.27 & \second{98.35} & 14.56 & 97.39 \\
KNN+ & 2.70 & 99.61 & 23.05 & 94.88 & 7.89 & 98.01 & 24.56 & 96.21 & 10.11 & 97.43 & 13.66 & 97.23 \\
CIDER & 2.89 & 99.72 & 23.88 & 94.09 & 5.75 & 99.01 & 20.21 & 96.64 & 12.33 & 96.85 & 13.01 & 97.26 \\
PALM & \best{0.34} & \best{99.91} & 28.81 & 94.80 & \best{1.11} & \best{99.65} & 34.07 & 95.17 & 10.48 & 98.29 & 14.96 & 97.57 \\
DPL & 1.51 & 99.71 & 19.65 & \second{96.39} & \second{2.24} & \second{99.47} & \second{12.41} & \second{97.84} & \second{7.23} & \best{98.81} & \best{8.61} & \best{98.45} \\
DRL & 7.91 & 98.82 & \second{19.17} & 95.65 & 12.87 & 99.09 & \best{11.92} & \best{98.12} & \best{4.92} & 97.48 & 11.58 & 97.83 \\
\midrule
\rowcolor{cyan!12}
\textbf{Ours} & \second{1.36} & \second{99.75} & \best{18.57} & \best{96.89} & 5.17 & 98.54 & 13.69 & 97.14 & 9.42 & 97.54 & \second{9.64} & \second{97.97} \\
\bottomrule
\end{tabular*}
\end{table*}

\begin{table*}[!t]
\centering
\caption{OOD detection results on CIFAR-100 with ResNet-34. FPR denotes FPR95. $\uparrow$ indicates larger values are better and $\downarrow$ indicates smaller values are better. Best results are shown in bold and second-best results are underlined.}
\label{table1}

\footnotesize
\setlength{\tabcolsep}{2.8pt}     % 横向压缩核心参数
\renewcommand{\arraystretch}{1.15}

\begin{tabular*}{\textwidth}{@{\extracolsep{\fill}}lcccccccccccc}
\toprule

& \multicolumn{10}{c}{\textbf{OOD Datasets}} 
& \multicolumn{2}{c}{\textbf{Average}} \\

\textbf{Methods}
& \multicolumn{2}{c}{SVHN}
& \multicolumn{2}{c}{Places365}
& \multicolumn{2}{c}{LSUN}
& \multicolumn{2}{c}{iSUN}
& \multicolumn{2}{c}{Textures}
& \\

\cmidrule(lr){2-13}

& FPR$\downarrow$ & AUROC$\uparrow$
& FPR$\downarrow$ & AUROC$\uparrow$
& FPR$\downarrow$ & AUROC$\uparrow$
& FPR$\downarrow$ & AUROC$\uparrow$
& FPR$\downarrow$ & AUROC$\uparrow$
& FPR$\downarrow$ & AUROC$\uparrow$ \\

\midrule

MSP & \tnum{78.89} & \tnum{79.80} & \tnum{84.38} & \tnum{74.21} & \tnum{83.47} & \tnum{75.28} & \tnum{84.61} & \tnum{74.51} & \tnum{86.51} & \tnum{72.53} & \tnum{83.97} & \tnum{75.27} \\
ODIN & \tnum{70.16} & \tnum{84.88} & \tnum{82.16} & \tnum{75.19} & \tnum{76.36} & \tnum{80.10} & \tnum{79.54} & \tnum{79.16} & \tnum{85.28} & \tnum{75.23} & \tnum{78.70} & \tnum{78.91} \\
Energy & \tnum{66.91} & \tnum{85.25} & \tnum{81.41} & \tnum{76.37} & \tnum{59.77} & \tnum{86.69} & \tnum{66.52} & \tnum{84.49} & \tnum{79.01} & \tnum{79.96} & \tnum{70.72} & \tnum{82.55} \\
ReAct & \tnum{50.93} & \tnum{88.75} & \tnum{83.55} & \tnum{73.10} & \tnum{64.02} & \tnum{80.31} & \tnum{81.80} & \tnum{79.99} & \tnum{64.40} & \tnum{81.95} & \tnum{68.94} & \tnum{80.82} \\
ASH & \tnum{52.96} & \tnum{90.19}   % SVHN
& \tnum{72.62} & \tnum{76.38}   % Places365
& \tnum{75.18} & \tnum{76.52}   % LSUN
& \tnum{55.55} & \tnum{87.86}   % iSUN
& \tnum{56.17} & \tnum{86.75}   % Textures
& \tnum{62.50} & \tnum{83.53}   % Average
\\
Vim
& \tnum{73.42} & \tnum{84.62}   % SVHN
& \tnum{85.34} & \tnum{69.34}   % Places365
& \tnum{86.96} & \tnum{69.74}   % LSUN
& \tnum{85.35} & \tnum{73.16}   % iSUN
& \tnum{74.56} & \tnum{76.23}   % Textures
& \tnum{81.13} & \tnum{74.62}   % Average
\\
VOS 
& \tnum{43.24} & \tnum{82.80}   % SVHN
& \tnum{76.85} & \tnum{78.63}   % Places365
& \tnum{73.61} & \tnum{84.69}   % LSUN
& \tnum{69.65} & \tnum{86.32}   % iSUN
& \tnum{57.57} & \tnum{87.31}   % Textures
& \tnum{64.18} & \tnum{83.95}   % Average
\\
SSD+ & \tnum{87.09} & \tnum{80.62} & \tnum{84.63} & \tnum{73.89} & \tnum{84.15} & \tnum{79.43} & \tnum{83.18} & \tnum{78.83} & \tnum{61.72} & \tnum{84.87} & \tnum{80.15} & \tnum{79.53} \\
KNN+ & \tnum{39.23} & \tnum{92.78} & \tnum{80.74} & \tnum{77.58} & \tnum{48.99} & \tnum{89.30} & \tnum{74.99} & \tnum{82.69} & \tnum{57.15} & \tnum{88.35} & \tnum{60.22} & \tnum{86.14} \\

CIDER & \tnum{23.09} & \tnum{95.16} & \tnum{79.63} & \tnum{73.43} & \tnum{16.16} & \tnum{96.33} & \tnum{71.68} & \tnum{82.98} & \tnum{43.87} & \tnum{90.42} & \tnum{46.89} & \tnum{87.66} \\

PALM 
& \best{\tnum{3.03}} & \best{\tnum{99.23}}
& \second{\tnum{67.80}} & \second{\tnum{82.62}}
& \best{\tnum{10.58}} & \best{\tnum{97.70}}
& \tnum{41.56} & \tnum{91.36}
& \tnum{44.06} & \tnum{91.43}
& \tnum{33.41} & \second{\tnum{92.47}} \\

DPL
& \tnum{14.44} & \tnum{98.25}   % SVHN
& \tnum{70.33} & \tnum{80.94}   % Places365
& \tnum{22.37} & \tnum{95.70}   % LSUN
& \second{\tnum{30.55}} & \second{\tnum{93.26}}   % iSUN
& \best{\tnum{26.82}} & \best{\tnum{93.05}}   % Textures
& \second{\tnum{32.90}} & \tnum{91.64}   % Average
\\
DRL
& \tnum{20.15} & \tnum{94.07}   % SVHN
& \tnum{76.64} & \tnum{77.55}   % Places365
& \tnum{16.97} & \tnum{94.63}   % LSUN
& \tnum{32.57} & \tnum{92.33}   % iSUN
& \second{\tnum{31.97}} & \second{\tnum{92.09}}   % Textures
& \tnum{35.66} & \tnum{90.13}   % Average
\\

\midrule

\rowcolor{cyan!12}
\textbf{Ours} 
& \second{\tnum{3.76}} & \second{\tnum{99.08}} 
& \best{\tnum{59.13}} & \best{\tnum{83.46}} 
& \second{\tnum{12.86}} & \second{\tnum{97.59}} 
& \best{\tnum{14.14}} & \best{\tnum{96.58}} 
& \tnum{38.78} & \tnum{90.38} 
& \best{\tnum{25.73}} & \best{\tnum{93.42}} \\

\bottomrule
\end{tabular*}
\end{table*}

\paragraph{Results on CIFAR.}
Tables~\ref{tab:cifar10_benchmark} and~\ref{table1} report CIFAR-10 and CIFAR-100 results using the same method list. EIHF is more effective on CIFAR-100, a setting with finer class structure and greater room for geometry-sensitive separation. It obtains the best average FPR95 and AUROC, improving the average FPR95 from 32.90 for DPL to 25.73 and the average AUROC from 92.47 for PALM to 93.42. The gains are most visible on Places365 and iSUN, while Textures remains a harder shift where DPL and DRL are stronger. On CIFAR-10, several baselines are already near saturation on multiple OOD sets, so EIHF gives a smaller but still competitive gain, reaching the second-best average FPR95 and AUROC.

The three-seed CIFAR-100 comparison in Table~\ref{tab:appendix_cifar100_multirun} checks whether this gain is tied to a single run; EIHF consistently lowers average FPR95 while maintaining comparable AUROC.

\paragraph{Results on ImageNet-100.}
Table~\ref{table2} evaluates ImageNet-100 with a ResNet-50 backbone. EIHF achieves the best FPR95 and AUROC on SUN and Textures, ranks second on iNaturalist, and obtains the best average FPR95 with an average AUROC close to DPL. Places is the main negative case: EIHF has the highest Places FPR95 among the compared methods and only matches DPL in AUROC. This result bounds the claim: early high-frequency evidence helps most for local-statistics or texture-heavy shifts, but is less reliable for scene-centric shifts that require stronger global semantics.

\begin{table*}[t]
\centering
\caption{OOD detection results on ImageNet-100 with ResNet-50. 
$\uparrow$ indicates larger values are better and $\downarrow$ indicates smaller values are better. 
Best results are shown in bold and second-best results are underlined.}
\label{table2}

\small
\setlength{\tabcolsep}{2.5pt}
\renewcommand{\arraystretch}{1.15}

\begin{tabular}{lcccccccccc}
\toprule

\textbf{Method}
& \multicolumn{2}{c}{\textbf{SUN}}
& \multicolumn{2}{c}{\textbf{Places}}
& \multicolumn{2}{c}{\textbf{Textures}}
& \multicolumn{2}{c}{\textbf{iNaturalist}}
& \multicolumn{2}{c}{\textbf{Average}} \\

& FPR95$\downarrow$ & AUROC$\uparrow$
& FPR95$\downarrow$ & AUROC$\uparrow$
& FPR95$\downarrow$ & AUROC$\uparrow$
& FPR95$\downarrow$ & AUROC$\uparrow$
& FPR95$\downarrow$ & AUROC$\uparrow$ \\

\midrule

SSD+  
& \second{30.34} & 93.06
& \best{34.38} & \second{91.52}
& 26.49 & 94.84
& 38.19 & 90.96
& 32.35 & 92.60 \\

KNN+  
& 41.85 & 92.25
& 44.41 & 90.26
& 26.60 & 94.42
& 38.54 & 94.15
& 37.85 & 92.72 \\

CIDER 
& 42.26 & 92.84
& 42.81 & 91.39
& 19.31 & 95.44
& 45.49 & 92.83
& 37.47 & 93.12 \\

PALM  
& 42.37 & 93.20
& \second{41.22} & \best{91.95}
& 17.02 & 96.16
& 32.08 & 95.14
& 33.17 & 94.11 \\

DPL  
& 38.70 & \second{93.95}
& 48.25 & 89.25
& \second{9.95}  & \second{98.97}
& \best{10.05} & \best{98.86}
& \second{26.74} & \best{95.26} \\
\midrule

\rowcolor{cyan!12}
Ours
& \best{23.83} & \best{95.52}
& 52.03 & 89.23
& \best{1.41}  & \best{99.58}
& \second{22.88} & \second{96.52}
& \best{25.04} & \second{95.21} \\

\bottomrule
\end{tabular}
\end{table*}

\begin{table*}[t]
\centering
\begin{minipage}[t]{0.55\textwidth}
\centering
\vspace{0pt}
\captionsetup{skip=2pt}
\caption{Scoring-protocol sensitivity on CIFAR-100 with ResNet-34. Baseline uses the corresponding RGB representation, while EIHF uses the four-channel representation; both are evaluated with the listed OOD scoring protocols.}
\label{tab:eihf_score_sensitivity}
\scriptsize
\setlength{\tabcolsep}{4pt}
\renewcommand{\arraystretch}{1.10}
\begin{tabular}{lcccc}
\toprule
\multirow{2}{*}{\textbf{Method}}
& \multicolumn{2}{c}{\textbf{Baseline}}
& \multicolumn{2}{c}{\textbf{EIHF}} \\
\cmidrule(lr){2-3}\cmidrule(lr){4-5}
& FPR$\downarrow$ & AUROC$\uparrow$
& FPR$\downarrow$ & AUROC$\uparrow$ \\
\midrule
MSP          & 83.97 & 75.27 & 80.90 & 73.95 \\
Energy       & \second{70.72} & \second{82.55} & 61.80 & 83.53 \\
SSD+         & 80.15 & 79.53 & \second{59.01} & \second{86.32} \\
PALM         & \best{33.41} & \best{92.47} & \best{25.73} & \best{93.42} \\
\bottomrule
\end{tabular}
\end{minipage}
\hfill
\begin{minipage}[t]{0.40\textwidth}
\centering
\vspace{0pt}
\captionsetup{skip=2pt}

\caption{Fourth-channel ablation on CIFAR-100 with ResNet-34. All variants use the same training and Mahalanobis scoring protocol; FPR95$\downarrow$, AUROC$\uparrow$.}
\label{tab:channel_ablation}

\vspace{0.25em}

\footnotesize
\setlength{\tabcolsep}{5pt}
\renewcommand{\arraystretch}{1.10}
\begin{tabular}{@{}lcc@{}}
\toprule
\textbf{Variant}
& \textbf{FPR95}$\downarrow$
& \textbf{AUROC}$\uparrow$ \\
\midrule
Zero
& \tnum{54.02}
& \tnum{84.82} \\
Random
& \tnum{35.88}
& \tnum{89.62} \\
Low-freq.
& \second{\tnum{28.03}}
& \second{\tnum{92.58}} \\
Shuffled HF
& \tnum{39.54}
& \tnum{90.02} \\
EIHF
& \best{\tnum{25.73}}
& \best{\tnum{93.42}} \\
\bottomrule
\end{tabular}
\end{minipage}
\end{table*}
\paragraph{Results on score sensitivity.}
Table~\ref{tab:eihf_score_sensitivity} tests whether the same EIHF representation helps all scoring protocols equally on CIFAR-100. The effect is score-dependent: EIHF improves Energy and SSD+, but the strongest gain is with Mahalanobis scoring, where average FPR95 drops from 33.41 to 25.73 and average AUROC rises from 92.47 to 93.42. MSP shows only a small FPR95 reduction with lower AUROC. These results support interpreting EIHF as a representation-side change whose benefit is strongest when the downstream score exploits class-conditional geometry.

\subsection{Ablation Studies}

\paragraph{What should the fourth channel contain?}
Table~\ref{tab:channel_ablation} tests what information the additional channel must carry. The zero- and random-channel controls isolate extra input capacity and stochastic regularization, but both remain well below EIHF on average. A low-frequency channel is the strongest non-EIHF variant, which shows that not every fourth channel is harmful; however, it still trails the aligned high-frequency residual in both average FPR95 and AUROC. The shuffled high-frequency variant keeps the marginal high-frequency statistics but loses pixel-level correspondence with the RGB image, and its performance drops accordingly. The gain therefore depends on local high-frequency structure aligned with the original image, not on the fourth channel itself.

\begin{table*}[t]
\centering
\vspace{0mm}

% ---------- left panel ----------
\begin{minipage}[t]{0.47\textwidth}
\centering
\vspace*{-32mm}

\begin{minipage}[t]{\linewidth}
\centering
\scriptsize
\setlength{\tabcolsep}{4.5pt}
\renewcommand{\arraystretch}{1.16}
\begin{tabular*}{0.92\linewidth}{@{\extracolsep{\fill}}lcc@{}}
\toprule
\textbf{Injection site} & \textbf{FPR95}$\downarrow$ & \textbf{AUROC}$\uparrow$ \\
\midrule
After stem   & 32.29 & 92.12 \\
After layer1 & 45.18 & 89.17 \\
After layer2 & 31.18 & 91.33 \\
\midrule
EIHF input   & \best{25.73} & \best{93.42} \\
\bottomrule
\end{tabular*}
\end{minipage}

\vspace{1.5mm}
{\footnotesize\textbf{(a)} Injection-location ablation\par}

\end{minipage}
\hfill
% ---------- right panel ----------
\begin{minipage}[t]{0.47\textwidth}
\centering

\begin{minipage}[t]{\linewidth}
\centering
\includegraphics[width=0.64\linewidth]{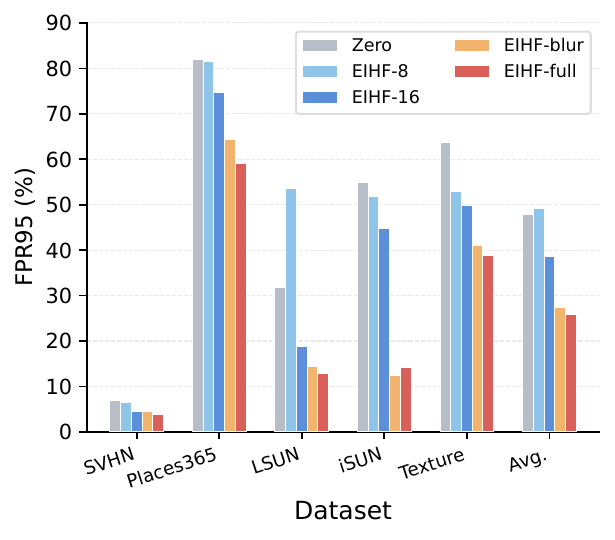}
\end{minipage}

\vspace{0.5mm}
{\footnotesize\textbf{(b)} High-frequency degradation ablation\par}

\end{minipage}

\vspace{0.75mm}
\caption{
Injection-site and residual-degradation ablations on CIFAR-100 with ResNet-34.
Panel (a) compares where the residual is injected; panel (b) degrades the residual spatially. FPR95$\downarrow$ is reported in both panels, and AUROC$\uparrow$ is additionally reported in (a).
}
\label{tab:injection_location}
\end{table*}

\paragraph{Does the injection location matter?}
Table~\ref{tab:injection_location}(a) compares EIHF with variants that inject the same Gaussian residual after the stem, layer1, or layer2 while keeping the backbone, PALM objective, scaling rule, and Mahalanobis scoring fixed. EIHF achieves the lowest average FPR95 and highest average AUROC, while later injection is less reliable. This pattern favors exposing the residual before the first convolution, where it can interact with RGB content from the earliest representation stage.

\paragraph{Does spatial preservation of the residual matter?}
Table~\ref{tab:injection_location}(b) compares the full-resolution residual with downsampled and blurred variants. Downsampling sharply weakens OOD performance, and the $8\times8$ variant can fall below the zero-channel control; blurring is less damaging. The result points away from pixel-level noise and toward spatially organized edge and texture cues.

\paragraph{Limitations.}
EIHF is not a universal improvement across OOD shifts or scores. Its gains concentrate in geometry-sensitive Mahalanobis-style detection and local-statistics shifts; scene-centric shifts remain challenging, as shown by weaker Places performance on ImageNet-100. Thus, early high-frequency injection complements rather than replaces semantic representation learning.

\section{Conclusion}

Band-wise \(\mathrm{MMD}^{2}\) suggests that mid- and high-frequency inputs induce stronger ID/OOD feature discrepancy than low-frequency inputs in our tested settings. EIHF tests this observation by appending a fixed high-frequency residual channel before the first convolution, improving geometry-sensitive OOD detection most clearly on CIFAR-100 and local-statistics shifts. The gains align with reduced Mahalanobis score overlap, making EIHF a representation-side intervention.

%%%%%%%%%%%%%%%%%%%%%%%%%%%%%%%%%%%%%%%%%%%%%%%%%%%%%%%%%%%%

\newpage
\appendix

\section{Additional CIFAR-100 Multi-Run Results}

\begin{table*}[h]
\centering
\caption{Three-seed stability comparison between PALM and EIHF on labeled CIFAR-100 using ResNet-34. Results are reported as mean$\pm$standard deviation over three independent runs. Best mean values are shown in bold.}
\label{tab:appendix_cifar100_multirun}
\small
\setlength{\tabcolsep}{8pt}
\renewcommand{\arraystretch}{1.12}
\begin{tabular}{lcc}
\toprule
\textbf{Method} & \textbf{Avg. FPR95}$\downarrow$ & \textbf{Avg. AUROC}$\uparrow$ \\
\midrule
PALM & $31.34{\pm}3.88$ & $93.02{\pm}1.08$ \\
EIHF & $\best{26.42}{\pm}0.73$ & $\best{93.08}{\pm}0.41$ \\
\bottomrule
\end{tabular}
\end{table*}

\paragraph{Stability across random seeds.}
To validate the stability of the matched CIFAR-100 comparison, we train PALM and
EIHF using three distinct random seeds and report the average and standard
deviation of FPR95 and AUROC in Table~\ref{tab:appendix_cifar100_multirun}.
Together with Table~\ref{table1} in the main paper, these results show that
PALM remains a strong baseline with small variation across independent runs.
EIHF shows stable average performance, consistently improving average FPR95
while maintaining comparable average AUROC under the same CIFAR-100 protocol.

\section{Additional ID Classification Accuracy}

\begin{figure}[h]
    \centering
    \includegraphics[width=0.62\linewidth]{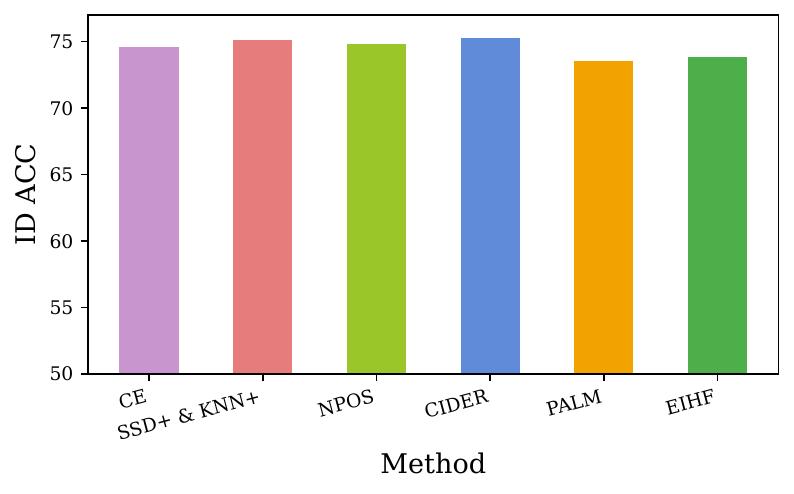}
    \caption{ID classification accuracy on CIFAR-100, comparing EIHF with the representation-learning and OOD baselines used in the main experiments.}
    \label{fig:appendix_id_acc}
\end{figure}

\paragraph{ID classification accuracy.}
We report the ID classification accuracy in Figure~\ref{fig:appendix_id_acc}.
Following the evaluation style of PALM~\citep{lu2024palm}, we compare standard
cross-entropy (CE), SSD+~\citep{sehwag2021ssd} and KNN+~\citep{sun2022knn}
under the same SupCon training objective~\citep{khosla2020supcon},
NPOS~\citep{tao2023npos},
CIDER~\citep{ming2023cider}, PALM, and EIHF. For our reproduced PALM and EIHF
models, we use the same CIFAR-100 ResNet-34 setting and projection-space
nearest-center evaluation protocol. PALM achieves \(73.54\%\) ID accuracy,
while EIHF achieves \(73.87\%\). These results show that EIHF maintains
comparable ID classification accuracy while improving OOD detection, indicating
that the proposed high-frequency injection does not compromise the ID
classification ability of the learned representation.

\section{Additional Geometry Diagnostics}

\begin{figure}[h]
    \centering
    \includegraphics[width=0.58\linewidth]{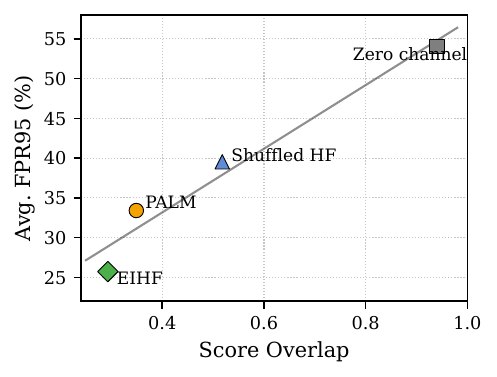}
    \caption{Mahalanobis score overlap versus average FPR95 on CIFAR-100. Lower overlap corresponds to less ID/OOD score mass shared near the decision region.}
    \label{fig:score_overlap_fpr}
\end{figure}

Table~\ref{tab:geometry_diagnostic} reports additional geometry diagnostics for the matched CIFAR-100 setting. 
Although the zero-channel control has the smallest within-class variance, it produces the largest Mahalanobis score overlap and the worst FPR95. 
In contrast, EIHF achieves the lowest score overlap and the best OOD performance. 
This shows that ID compactness alone is not sufficient to explain OOD detection performance.
Figure~\ref{fig:score_overlap_fpr} further visualizes the relationship between score overlap and average FPR95. 
Across the compared variants, lower Mahalanobis score overlap generally corresponds to lower FPR95, supporting our interpretation that EIHF mainly improves Mahalanobis-style detection by reducing ID/OOD score overlap rather than merely shrinking ID features.

\begin{table*}[h]
\centering
\caption{Geometry diagnostics on CIFAR-100 with ResNet-34. Score overlap is the
normalized histogram overlap between ID and OOD Mahalanobis scores, averaged
over all OOD samples. Best results are shown in bold and second-best results are underlined.}
\label{tab:geometry_diagnostic}
\small
\setlength{\tabcolsep}{6pt}
\begin{tabular}{lccccc}
\toprule
\textbf{Variant}
& \(\mathbf{V}_{\mathrm{intra}}\downarrow\)
& \textbf{Mean ID Dist.}\(\downarrow\)
& \textbf{Score Overlap}\(\downarrow\)
& \textbf{Avg. FPR95}\(\downarrow\)
& \textbf{Avg. AUROC}\(\uparrow\) \\
\midrule
PALM & 0.434 & 0.588 & \second{0.349} & \second{33.41} & \second{92.47} \\
Zero channel & \best{0.348} & \second{0.572} & 0.940 & 54.02 & 84.82 \\
Shuffled high-frequency & \second{0.394} & 0.590 & 0.518 & 39.54 & 90.02 \\

EIHF & 0.424 & \best{0.465} & \best{0.293} & \best{25.73} & \best{93.42} \\
\bottomrule
\end{tabular}
\end{table*}

\section{Additional High-Frequency Operator Ablation}

\begin{table}[h]
\centering
\caption{High-frequency operator ablation on CIFAR-100 with ResNet-34 under the same controlled protocol.}
\label{tab:hf_operator_ablation}
\small
\setlength{\tabcolsep}{6pt}
\renewcommand{\arraystretch}{1.12}
\begin{tabular}{lcc}
\toprule
\textbf{Operator} & \textbf{Avg. FPR95}$\downarrow$ & \textbf{Avg. AUROC}$\uparrow$ \\
\midrule
FFT      & 40.07 & 88.54 \\
Laplace  & 38.23 & 90.31 \\
Sobel    & \second{31.22} & \best{91.92} \\
Gaussian & \best{29.41} & \second{90.89} \\
\bottomrule
\end{tabular}
\end{table}

\paragraph{High-frequency operator sensitivity.}
We test whether EIHF depends on a particular high-pass operator under a controlled protocol where all variants share the same backbone, training objective, batch size, scaling rule, and Mahalanobis scoring function. Table~\ref{tab:hf_operator_ablation} is therefore a within-study comparison rather than a direct re-ranking against the main EIHF result. Gaussian residual gives the best average FPR95, while Sobel gives the highest average AUROC. Since FPR95 is our primary OOD metric, we use Gaussian residual as the default operator; it also produces an isotropic residual and avoids the direction-specific bias of gradient filters.

\section{Additional Diagnostic Details}

For the band-wise \(\mathrm{MMD}^{2}\) diagnostic, all comparisons use fixed encoders trained before the diagnostic is computed. The Fourier masks partition the normalized radial frequency plane into non-overlapping annular bands, and each band-limited input is passed through the same preprocessing and final-feature extractor as the corresponding full-image model. We use the same ID/OOD sampling protocol for every band within a comparison, and the RBF kernel bandwidth is fixed before comparing the resulting band-wise curves. These choices make the diagnostic a relative frequency-profile comparison rather than a tuned OOD detector.

\section{Implementation Details for EIHF Preprocessing}

For PALM training, each view is first generated by the standard SimCLR-style
transformation, including random resized crop, horizontal flip, color jitter,
random grayscale conversion, tensor conversion, and normalization. For two-crop
training, the two views are concatenated along the batch dimension before EIHF is
applied. The residual channel is then computed from the normalized RGB tensor
using a fixed grayscale conversion followed by a Gaussian residual:
\[
G(x)=0.2989x_R+0.5870x_G+0.1140x_B,\quad
C_{\mathrm{hf}}(x)=|G(x)-K*G(x)|.
\]
The scaled residual \(e(x)=\alpha_{\mathrm{hf}}C_{\mathrm{hf}}(x)\) is
concatenated with the normalized RGB tensor. The same order is used at
evaluation time after deterministic resize/crop and normalization.

%%%%%%%%%%%%%%%%%%%%%%%%%%%%%%%%%%%%%%%%%%%%%%%%%%%%%%%%%%%%

\newpage
% Checklist removed for arXiv package.

\end{document}